\documentclass[manuscript]{acmart}
\usepackage{endnotes}

\usepackage{dcolumn}
\usepackage{xcolor}
\usepackage{csquotes}
\usepackage{subcaption}
\usepackage{multicol}

\usepackage{booktabs}
\usepackage{graphicx}

\newcommand{\xhdr}[1]{\vspace{1mm} \noindent\textbf{#1}}
\usepackage[subtle]{savetrees}

\usepackage{color-edits}
\addauthor[Emily]{eb}{blue}
\addauthor[Hoda]{hh}{red}
\addauthor[Kit]{kr}{brown}
\addauthor[Rayid]{rg}{purple}
\addauthor[Rocket]{rn}{magenta}

\newcolumntype{d}{D{.}{.}{-1}}

\AtBeginDocument{%
  \providecommand\BibTeX{{%
    \normalfont B\kern-0.5em{\scshape i\kern-0.25em b}\kern-0.8em\TeX}}}

\copyrightyear{2023} 
\acmYear{2023} 
\setcopyright{rightsretained} 
\acmConference[EAAMO '23]{Equity and Access in Algorithms, Mechanisms, and Optimization}{October 30-November 1, 2023}{Boston, MA, USA}
\acmBooktitle{Equity and Access in Algorithms, Mechanisms, and Optimization (EAAMO '23), October 30-November 1, 2023, Boston, MA, USA}
\acmDOI{10.1145/3617694.3623259}
\acmISBN{979-8-4007-0381-2/23/11}
\acmConference[EAAMO '23]{ACM Conference on Equity and Access in Algorithms, Mechanisms, and Optimization '23}{October 30 - November 1, 2023}{Boston, USA}
\acmBooktitle{ACM Conference on Equity and Access in Algorithms, Mechanisms, and Optimization 2023}

\begin{document}



\title{Toward Operationalizing Pipeline-aware ML Fairness: \\A Research Agenda for Developing Practical Guidelines and Tools}

\author{Emily Black}
\email{eblack@barnard.edu}
\affiliation{%
  \institution{Barnard College, Columbia University}
  \country{USA}
 }

\author{Rakshit Naidu}
\email{rakshitnaidu@gatech.edu}
\affiliation{%
  \institution{Georgia Institute of Technology}
  \country{USA}
}


\author{Rayid Ghani}
\email{rayid@cmu.edu}
\affiliation{%
  \institution{Carnegie Mellon University}
  \country{USA}
}

\author{Kit T. Rodolfa}
\email{krodolfa@law.stanford.edu}
\affiliation{%
  \institution{Stanford University}
  \country{USA}
}

\author{Daniel E. Ho}
\email{dho@law.stanford.edu}
\affiliation{%
  \institution{Stanford University}
  \country{USA}
  }

\author{Hoda Heidari}
\email{hheidari@cmu.edu}
\affiliation{%
  \institution{Carnegie Mellon University}
  \country{USA}
}

\newcommand{\w}{\mathbf{w}}

\renewcommand{\a}{\mathbf{a}}
\renewcommand{\w}{\mathbf{w}}

\renewcommand{\shortauthors}{Black et al.}

\begin{abstract}
While algorithmic fairness is a thriving area of research, in practice, mitigating issues of bias often gets reduced to enforcing an arbitrarily chosen fairness metric, either by enforcing fairness constraints during the optimization step, post-processing model outputs, or by manipulating the training data.
Recent work has called on the ML community to take a more holistic approach to tackle fairness issues by systematically investigating the many design choices made through the ML pipeline, and identifying interventions that target the issue's root cause, as opposed to its symptoms. While we share the conviction that this pipeline-based approach is the most appropriate for combating algorithmic unfairness on the ground, we believe there are currently very few methods of \emph{operationalizing} this approach in practice.  
Drawing on our experience as educators and practitioners, we first demonstrate that without clear guidelines and toolkits, even individuals with specialized ML knowledge find it challenging to hypothesize how various design choices influence model behavior.
We then consult the fair-ML literature to understand the progress to date toward operationalizing the pipeline-aware approach: we systematically collect and organize the prior work that attempts to detect, measure, and mitigate various sources of unfairness through the ML pipeline. We utilize this extensive categorization of previous contributions to sketch a research agenda for the community. We hope this work serves as the stepping stone toward a more comprehensive set of resources for ML researchers, practitioners, and students interested in exploring, designing, and testing pipeline-oriented approaches to algorithmic fairness.
\end{abstract}

\maketitle 

\section{Introduction} 

\begin{figure*}
    \centering
    
    \includegraphics[width=\textwidth]{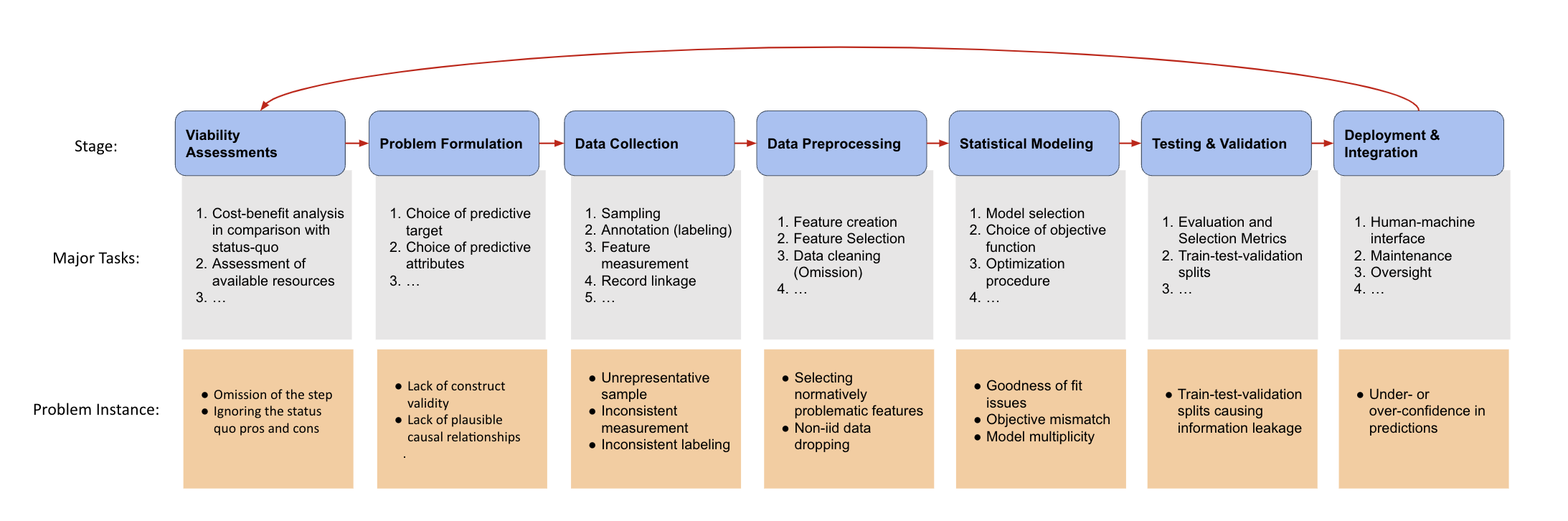}
    
    \caption{A simplified view of the ML pipeline, its key stages, and instances of design choices made at each stage.}
    \label{fig:pipeline}
\end{figure*}

As a result of rising public and legal pressure on technology companies and governments to create more equitable systems\cite{house2022blueprint,room2021executive,ftcproposed,ai2023artificial}, professionals across industry, government, and nonprofits have been turning to algorithmic fairness expertise to guide their implementation of AI systems~\cite{holstein2019improving}. 
While there has been extensive work in the service of preventing algorithmic unfairness~\cite{pessach2022review,agarwal2018reductions,hardt2016equality,petersenneurips21,kamiran2012data}, mitigation in practice often gets reduced to enforcing a somewhat arbitrarily formulated fairness metric on top of a pre-developed or deployed system~\cite{madaio2020co}. 
Practitioners often make ad-hoc mitigation choices to improve fairness metrics, such as removing sensitive attributes, changing the data distribution, 
enforcing fairness constraints, or post-processing model predictions. While these techniques may improve selected fairness metrics, they often have little practical impact; at worst, they can even exacerbate the same disparity metrics they aim to alleviate~\cite{wang2021fair,li2022more}. 
Prior scholarship has attributed these problematic trends to the fact that the traditional approach to fairness fails to take a system-wide view of the problem. They narrow bias mitigation to a restricted set of points along the ML pipeline (e.g., the choice of optimization function). This is despite the well-established fact that \emph{numerous} choices there can impact the model's behavior~\citep{suresh2021framework}. Assessing viability/functionality of AI~\citep{raji2022fallacy}, problem formulation~\cite{passi2019problem}, data collection, data pre-processing, 
modeling, testing and validation, and organizational integration are all key stages of the ML pipeline consisting of consequential design choices (see Figure~\ref{fig:pipeline} for an overview).
By abstracting away the ML pipeline and selecting an ad-hoc mitigation strategy, the mainstream approach misses the opportunity to identify, isolate, and mitigate the underlying \emph{sources} of unfairness, which can in turn lead to fairness-accuracy tradeoffs due to intervening at the wrong place~\cite{black2022algorithmic},  
or even worse,``[hide] the real problem while creating an illusion of
fairness''~\cite{akpinar2022sandbox}. 
\newline
\xhdr{The pipeline-aware approach to algorithmic fairness.}
We join prior calls advocating for an alternate \emph{pipeline-aware} approach to fairness~\cite{akpinar2022sandbox,suresh2021framework}.
At a high level, this approach works as follows: given a model with undesirable fairness behavior, the ML team must search for ways in which the variety of choices made across the ML creation pipeline may have contributed to the behavior (e.g., the choice of prediction target~\cite{obermeyer2019dissecting}). Once plausible causes are identified, the team should evaluate whether other choices could abate the problem (e.g. changing a model's prediction target and re-training~\cite{black2022algorithmic}).\footnote{This description is taking an auditing perspective: when building a fair model from scratch, fairness desiderata would be described, and the practitioners would enumerate choices that can be made at each step of the ML creation pipeline and avoid choices that work against this desired behavior.} This process should take place iteratively, and the model should be re-evaluated until it is deemed satisfactory, whereupon bias testing and model updates would continue throughout deployment.

\xhdr{The need for operationalizing the pipeline-aware approach} While prior work has clearly established the benefits of the pipeline-aware view toward fairness, we contend that \emph{conceptual awareness} of this approach alone won't be sufficient for \emph{operationalizing} it in practice. In Section~\ref{sec:challeneges}, we provide evidence suggesting that making informed hypotheses about the root causes of unfairness in the ML pipeline is a challenging task, even for individuals with specialized ML knowledge and skills. For example, based on qualitative data gathered from a graduate-level fair-ML class at an R1 institution, students with significant ML background struggle to conceptualize sources of harmful model behavior and suggest appropriately chosen bias mitigation strategies. This observation is, indeed, consistent with our experience working with ML practitioners and system developers across a wide range of public policy settings.
This evidence motivates our focus on ``operationalization'':
to \emph{find} less discriminatory models, we argue that practitioners need usable tools to measure the discrimination from, identify the underlying sources of, unfairness in the pipeline, and then match those underlying sources with appropriately designed interventions. 
Such tools would be instrumental in taking advantage of the legal systems already in place to reduce discrimination---such as the disparate impact doctrine, especially as they may ease legal concern over the direct use of protected attributes during training and deployment~\cite{colfax2021report1,colfax2021report2}, as we discuss in Section~\ref{sec:regulation}. 



\begin{figure*}
    \centering
    \includegraphics[width=\textwidth]{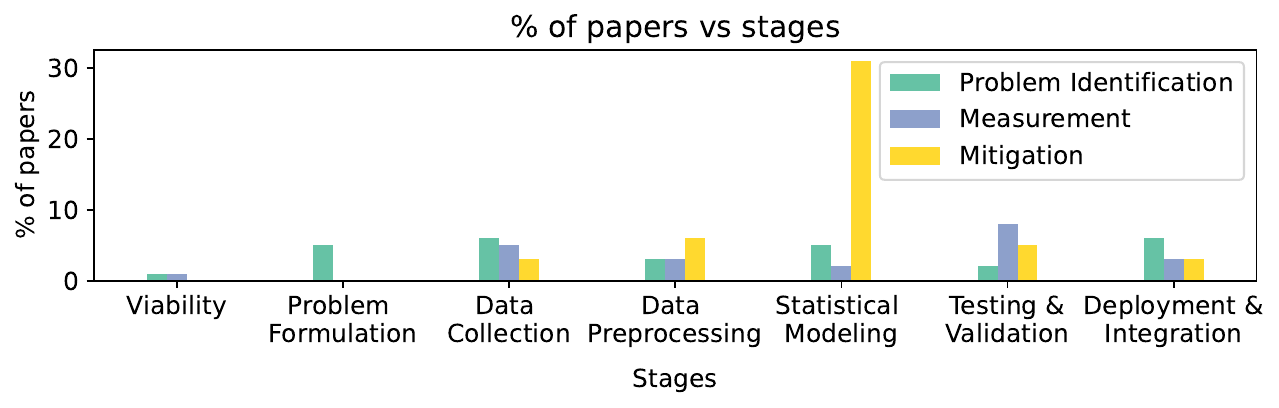}
    \vspace{-2.5em}
    \caption{The distribution of research efforts dedicated to different stages of the ML pipeline among the papers we surveyed.}
    \label{fig:distribution}
\end{figure*}

\xhdr{A snapshot of progress toward operationalization.} Having established the need for practical implementation of the pipeline-aware approach to fairness, we seek to understand the progress made so far, and ask: how far along are we toward operationalizing the pipeline-aware approach? Do we currently have useful methods and guidelines to inspect and modify the variety of design choices made throughout the ML pipeline in practice? 
To respond to these questions, we survey the Fair-ML literature
in search of methods that identify, measure, or mitigate biases arising from specific ML design choices, 
and map them to specific stages of the pipeline (Section~\ref{sec:litreview}). While we identify numerous gaps in the existing arsenal of tools, we hope our work offers practitioners a one-stop shop for identifying potential causes of unfairness in their use cases and getting an overview of the state of the art to detect, measure, and mitigate those issues. 
In service of this goal, we have turned our survey into an interactive and community-maintained wiki documenting pipeline-aware de-biasing tools for Fair-ML researchers, practitioners, and students interested in exploring, designing, and testing system-level approaches to algorithmic fairness.\footnote{http://fairpipe.dssg.io/}

\xhdr{Sketching a path forward for the research community.} Building off of our survey, in Section ~\ref{sec:agenda} 
outline a research agenda towards operationalizing the pipeline-aware approach, which we expand upon in Appendix 2.
As a preview of one of our insights, Figure~\ref{fig:distribution} depicts the relative effort the research community has allocated to different stages of the ML pipeline. As evident in this picture, the focus of the ML community has been largely on the statistical modeling stage, with an out-sized emphasis on mitigation strategies. 
This finding generalizes recent observations that ``Everyone wants to do the model work, not the data work''~\citep{sambasivan2021everyone}. 
Key stages of the pipeline, including viability assessment, problem formulation, and deployment and monitoring, have been understudied, as well as the interactions between them---we hypothesize due to lack of potential for novel, quantitative contributions or theoretical analysis~\citep{birhane2021values}. Beyond these and other gaps, we observe a disconnect between identifying problems in the pipeline, largely done by the HCI community, and mitigating them, largely done in the algorithmic fairness community, and that developing  guidelines to weigh the imperfect choices practitioners face against each other is a crucial avenue for collaboration between ML experts and ethicists.

Finally, we close by introducing a first version of a tool that ML practitioners can use to build AI systems with a pipeline-aware approach to fairness: pipeline cards, which we expand upon in Appendix 3. We hope that pipeline cards provide researchers and model practitioners with actionable step towards our using a pipeline-aware approach to fairness in practice.
\section{The Need to Operationalize the Pipeline-Aware Approach}
\label{sec:challeneges}
In this section, we highlight the need for, and difficulty in, \emph{operationalizing} a pipeline-aware approach to fairness. We first analyze a recent case study in anti-discrimination law to show how operationalizing a pipeline-aware approach to fairness is essential to take advantage of legal tools such as the disparate impact framework--especially as existing mitigation techniques run the risk of violating existing legal restrictions on the use of protected attributes. Following this, we present evidence from a classroom study where ML students struggle to correctly hypothesize about the underlying causes of unfairness and suggest plausible remedies--showing that conceptual understanding of the pipeline and the variety of choices within it are by no means sufficient to inform good practice. 

\subsection{Evidence from Regulatory Enforcement}\label{sec:regulation}
Recent developments in attempts to regulate the design and use of  ML systems have given a sense of urgency to support regulators and policymakers in these efforts~\cite{ai2023artificial,house2022blueprint,room2021executive,ftcANPR2023}. 
In this section, we focus on one of the first examples of adherence of anti-discrimination law in AI systems~\cite{colfax2021report1} to show how a pipeline-aware approach may be particularly helpful in establishing legal liability in, and providing remedies for, discrimination in regulated AI systems. 

In particular, we note that when searching for less discriminatory alternatives to deployed models, a pipeline-aware approach may elide some of the potential legal problems that apply to using more traditional algorithmic fairness approaches~\cite{ho2020affirmative, bent2019algorithmic}. Many traditional algorithmic fairness techniques that intervene at the modeling stage often use protected attributes to change model behavior by learning new prediction thresholds or modifying the training procedure~\cite{agarwal2018reductions,hardt2016equality}, leading regulators to dismiss such approaches due to legal restrictions around the use of protected attributes in decision-making~\cite{ho2020affirmative, bent2019algorithmic}. However, pipeline-aware techniques may use protected attribute labels to \emph{evaluate} different models and choose among them, but may not directly enforce a constraint using protected attributes, avoiding the same legal scrutiny applied to other algorithmic fairness techniques.
Thus, it is imperative that we build pipeline-aware tools to empower regulators, policymakers, and advocacy groups pushing for legal requirements surrounding bias reduction in public-facing AI systems to find, and enforce the use of, less discriminatory systems.

\xhdr{Case Study Background: Upstart Monitorship.} In 2020, the consumer finance firm Upstart, the NAACP, and Relman Colfax, a civil rights law firm, entered a legal agreement to investigate racial discrimination in Upstart's lending model due to the NAACP's concerns over the use of attributes related to educational attainment, such as the name of the college applicants attended if they had a college degree~\cite{naacp_statement}. The goal was  first to determine if there was a legally relevant difference in selection rate between Black and white applicants using Upstart's model, which they found there was~\cite{colfax2021report2}. Once this was established, Relman Colfax followed the disparate impact doctrine~\cite{disp_impact_doctrine} to determine whether Upstart would be legally required to change their model: to do so, they search for a "less discriminatory alternative" model, or LDA, with equivalent performance but less disparate impact across racial groups. Under the disparate impact law, once discrimination is established and evidence establishes an LDA, a company may be required to replace the discriminatory model with the LDA~\cite{disp_impact_doctrine}.

\xhdr{Searching for a Less Discriminatory Alternatives: Suboptimal Approaches.} 
After establishing discriminatory model behavior, third-party investigators developed and implemented a strategy to find a less discriminatory model with similar predictive performance to Upstart's original algorithm. Notably, the third-party bias investigators discounted all algorithmic fairness techniques to mitigate discrimination out of hand, seemingly from concern over legal repercussions over the use of the protected attribute to influence model behavior, and a desire to ``align with traditional principles gleaned from antidiscrimination
jurisprudence''~\cite{colfax2021report1}. As the authors of the bias investigation report note:
\begin{displayquote}
A range of techniques for mitigating disparities is proposed in the algorithmic fairness literature. \textit{Some of these proposals could raise independent fair lending risks, such as the use of different
models for different protected classes or the improper use of prohibited bases as predictive variables}...While [the algorithmic fairness]
conversation is valuable, many “fairness” proposals do not engage or align with the established three-step disparate
impact analysis reflected in case law and regulatory materials.~\cite{colfax2021report1}
\end{displayquote}
The procedure that was taken instead was intervening at the feature selection step, and searching the feature space for a model with a subset of features that was less discriminatory, drawing from well-established literature on differential item functioning~\cite{chen2014differential}. That is, the practitioners created a model for several different feature combinations, and tested the disparate impact of each one~\cite{colfax2021report2}. While a less discriminatory alternative was discovered,\footnote{The model discovered was less discriminatory but also suffered performance drop, which Relman Colfax argued was within an acceptable range to be an ``equivalent performance'', but the exact rules around this have yet to be established---so it is unclear if this model would in fact suffice as an LDA~\cite{colfax2022report3}.} this procedure took sufficiently long that Upstart updated its model before the investigations were completed~\cite{colfax2022report3}. We note that the blind, exhaustive search for a less discriminatory model at just one intervention point in the machine learning pipeline is almost certainly expensive, inefficient, and leads to suboptimal outcomes. If pipeline-aware tools had been available to better isolate sources of bias in the \textit{entire} pipeline, beyond feature choice, it is possible they would have found a less discriminatory model with acceptable predictive performance more efficiently.
 
The disparate impact framework gives advocates and model practitioners a way to challenge the use of discriminatory algorithms, and further incentivize companies to thoroughly explore possible algorithms to find the least discriminatory one within those with sufficient business utility---but using ineffective methods to enforce these regulatory tools weaken their power. We suggest that in order to effectively leverage the disparate impact doctrine, we must operationalize a pipeline-aware approach to ML fairness. And, even beyond searching for LDA models, tools informed by a pipeline-based approach may also aid in creating more standard and rigorous approaches to algorithmic audits more broadly.

\subsection{Evidence from the Classroom}\label{sec:teaching_evidence}
Our team has documented the challenges of instilling the pipeline-centric view of ML harms in students using traditional teaching methods (e.g., lectures containing real-world examples). The classroom activity outlined in this section was conducted by one of us at an R1 educational institution as part of a graduate-level course focused on the ethical and societal considerations around the use of ML in socially high-stakes domains. Our IRB approved the activity, and students had the option of opting out of data collection for research purposes.
\footnote{CMU's IRB approved the study: STUDY2022\_00000447}

\xhdr{Study Population and Design} We leave the majority of the details about study design and population to Appendix 1. However, we note that all 37 participants had at least one  prior class in Machine Learning, so they had non-negligible background. In the study, students were introduced to the ML pipeline through an approximately 45-minutes long lecture, with multiple examples of how design choices at each stage can lead to harmful outcomes, such as unfairness.
Next, they were then asked to team up with 3-4 classmates and pick a societal domain as the focus of their group activity, the domains they chose are outlined in Table 1. Third, students were given 30 minutes to discuss the following questions about their application domain with teammates and submit their written responses individually: \newline 1) Characterizing the specific \textbf{predictive task} their team focused on; 2) The \textbf{type of harm} observed; 3) Their \textbf{hypotheses around the sources} of this harm in through the ML pipeline; and 4) Their \textbf{hypotheses around potentially effective remedies} for addressing those sources.

\xhdr{Findings. }
A thematic analysis of submitted responses revealed several challenges: 

\xhdr{Theme 1: Specifying how a real-world problem gets translated into a predictive task was not straightforward.} Table 2 overviews how each team defined their predictive tasks. For example, the finance team defined the task as \textit{assessing the \emph{creditworthiness} of individuals using their demographic and socio-economic data}. The public safety team defined the predictive task as \textit{allocating police presence to \emph{high-risk} areas}. The social media team defined the task as \textit{predicting whether a news article is fake}. Note that notions of applicant's \emph{creditworthiness} or neighborhood's safety \emph{risk}, or \emph{fakeness} of an article's content are not well-defined targets. Other teams did not adequately distinguish between the construct of interest and its operationalization as the target of prediction. For example, some members of the organ transplant team characterized the task as deciding \textit{which people \emph{should} receive an organ transplant}. In contrast, others characterized it as \textit{predicting whether an individual would receive an organ transplant in a given hospital}. Note the difference between \emph{``should''} and \emph{``would''}.

\xhdr{Theme 2: harms were characterized in broad strokes.}
Some teams described harm without specifying the groups or communities that could be impacted by it and the baseline of comparison. For example, the housing team stated \textit{price discrepancies} due to property location as the harm occurring in their domain but did not specify who could be negatively impacted by price discrepancies and in comparison with what reference group this should be considered a harm. Another example was \textit{spread of fake news} as the harm without mentioning whom it can impact negatively and how.

\xhdr{Theme 3: Students had difficulty mapping the observed harm to a plausible underlying cause.} For example, the child welfare group attempted to explain the harm against Black communities by noting that the feature, Race, was not quantified with sufficient granularity. The tool allowed only three racial categories: White, Black, and Other, and they hypothesized that that could be the cause of the disparity. Note that there is no plausible mechanism through which this lack of granularity might have led to disparities in child welfare risk assessments against Black communities. As another example, the finance team offered \textit{biases of developers} as the potential source of disparity in lending practices.

\xhdr{Theme 4: Students had difficulty mapping their hypothesized causes to plausible remedies.} For example, the Child Welfare team proposed randomly selecting instances for inclusion in the training data. The online exam proctoring group suggested further transparency (e.g., telling students what behavior results in a cheating flag) to reduce errors. 

\xhdr{Theme 5: Students used broad-stroke language to describe causes and remedies of harm.} For example, several teams referred to \textit{biased data} as the underlying cause and offered \textbf{more comprehensive data collection} as the remedy. While correct, such high-level assessments are unlikely to lead to concrete actions in practice. Another commonly proposal was \textit{human oversight} of decisions without specifying the ramifications of leaving the final call to human decision-makers.

In a session after the data collection and analysis, the instructor led a class-wide discussion in which students were encouraged to reflect on the activity and some of the gaps in the arguments presented by student teams.

\section{A Survey of Fairness Research Along the Pipeline}
\label{sec:litreview}

As a first step towards operationalizing the pipeline-based approach to fairness in ML, we provide a review of the ML fairness literature focused on creating a taxonomy of fairness work that locates, measures, or mitigates problems along the ML pipeline. In addition to serving as 
a resource that ML practitioners can use to identify ways to diagnose and mitigate problems in their pipeline, as we expand on in the Section~\ref{sec:agenda}, our survey of the work already done allows us to point to the gaps in the literature that must be addressed to have a full understanding of how choices made along the ML pipeline translate to model behavior--- and create tools which operationalize this understanding to build more effective systems.

\subsection{Survey Methods and Organization.}
To better understand the landscape of algorithmic fairness research throughout the ML pipeline, we performed a thorough survey of the recent literature, which we classified depending upon which area of the pipeline analyzed. We gathered papers from NEURIPS, ICML, ICLR, FAccT, AIES, EAAMO, CHI, and CSCW for the past five years, i.e. 2018-2022, containing any of the following terms in their title, abstract, or keywords: "fairness", "fair", "discrimination", "disparity", "equity".\footnote{When available: this is with the exception of NEURIPS, which we gather for years 2017-2021 due to the date of the conference relative to the time writing this work, and EAAMO, which started in 2021.} In addition, we performed a series of Google Scholar searches to ensure our survey did not miss high-impact work published in other venues: One search used the keywords listed above and included papers published in any venue in the past five years with over 50 citations, through the top 50 results returned by this Google Scholar search. Additional Google Scholar searches used keywords from each step in the pipeline individually, to attempt to find papers targeted at each stage: for example, "data collection" and "fairness" and "machine learning". This results in \textasciitilde1000 papers overall which fit our search criteria.

We then manually inspect each paper to understand whether and how the reported research is related to the machine learning pipeline. 
That is, does it identify, measure, or mitigate a concrete cause of unfairness due to choices made in a specific stage of the ML pipeline. If so, we categorize the paper along two axes: what part of the pipeline it corresponds to (problem formulation, data choice, feature engineering, statistical modeling, testing and validation, or organizational realities), and whether it identifies, measures, mitigates, or provides a case study of a pipeline-based fairness problem.\footnote{By \textit{identifying} a pipeline fairness problem, we refer to papers that point to a previously unobserved source of bias on the machine learning pipeline either through theory or through experimentation with training pipelines on common machine learning datasets; by \textit{measuring} a pipeline fairness problem, we refer to papers that provide a generalized technique for how to identify or gauge the magnitude of a specific source of unfairness along the machine learning pipeline; by mitigating, we mean paper which develop a technique for addressing a source of bias along the AI pipeline when it arises; and by a case study we refer to an example of how a choice on the machine learning pipeline lead to unfairness in a specific application, often on an already deployed model.} Of our approximately 1000 papers, \textasciitilde300 satisfied our criteria of 
studying some aspect of the pipeline.

We present our full categorization of all the papers that we found related to 
the machine learning pipeline,
broken down into what stage of the pipeline they were most related to, and whether they were case study, identification, measurement, or mitigation papers, in Table 1 in the appendix. In our survey below, we give a sample of the space: we do not aim to be completely comprehensive, but instead, we aim to both highlight both some of the most well-known papers connected to each component of the pipeline, as well as those that offer a novel or promising 
perspective to understudied areas.

\subsection{Viability Assessments}
\subsubsection{Definition and Decisions}
Viability assessments refer to a series of early investigations into whether including an ML component within the decision-making system is preferable to the status quo of decision-making; and if so, if it is \emph{possible} to build a net-beneficial ML system given available resources including data, expertise, budget, and other organizational constraints. Examples of decisions in this stage include: 
What are the policy goals of the decision-making problem? How can introducing an ML model promote that goal? How should the ML component be scoped? 
Do we have 
stakeholder buy-in? 
Is there organizational capacity to build and maintain this algorithm? 

\subsubsection{Case Studies and Problem Identification} 
To the authors' knowledge, there is very little work within the fairness literature on documenting, or detailing the bias that can arise out of, or mitigating bias from the viability assessment process. \citet{raji2022fallacy} provide extensive evidence on numerous deployed algorithmic products that simply do not work--examples of badly scoped projects~\citep{oakden2019hidden,szalavitz2021pain,freeman2021use}. They warn against the presumption of AI functionality, and point to several failure modes that could be remedied with a viability assessment step: such as attempting conceptually or practically impossible tasks. \citet{wang2022against} point out several common flaws of predictive optimization, including the discrepancy between intervention vs. prediction, lack of construct validity, distribution shifts, and lack of contestability.
Viability assessments also provide an avenue for refusal to build an ML system---though there have been discussions in the community around when to refuse to build~\cite{barocas2020not}, there is little published work on case studies detailing how the decision to build or not build was made. Indeed, recent work has pointed to how organizational factors---such as lack of a company's support for ethical AI approaches, or time pressure, or focusing only on client demands~\cite{passi2020making} can lead to pushing ML systems ahead without careful consideration of whether or not deploying a system in that domain is a net benefit in the given context. 

\subsubsection{Measurement, Mitigation, and Tools} ML practitioners have proposed guidelines for initial scoping of ML projects\footnote{See, e.g., \url{http://www.datasciencepublicpolicy.org/our-work/tools-guides/data-science-project-scoping-guide/}}. More recently, \citet{wang2022against} have presented a rubric for assessing the \emph{legitimacy} of
predictive optimization. 
To our knowledge, existing proposals, while promising, have not yet been evaluated in practice.

\subsection{Problem Formulation}

\subsubsection{Definition and Decisions}
Problem formulation~\cite{passi2019problem} is the translation of a real-world problem to a prediction task: for example, turning a bank’s need to select certain individuals to give loans to,
into a machine learning system with numerical inputs and outputs. We focus on three main decisions here: selecting a  prediction target, what inputs should be used to predict that target, and the prediction universe. The selection of a prediction target translates the ultimate goal of a business or policy problem into a numerical data representation of that goal: for example, predicting “creditworthiness” by predicting the probability of missing a payment on a loan. Selection of inputs includes discussion over what features in the available data are acceptable to use to predict the target: e.g. whether it is normatively acceptable to use access to a telephone as a predictor of a failure to appear in pretrial risk assessment instruments~\cite{grgic2018beyond, lum2019measures}. The selection of the prediction universe determines who predictions are made over to solve this problem: for example, when predicting likelihood of not graduating high school, is this predicted over 7th graders, 9th graders, or 11th graders?
\subsubsection{Discussions and Considerations around Problem Formulation}
Several works have discussed what constitutes problem formulation~\cite{passi2019problem}, how the process occurs and can be influenced by organizational biases, and what factors are important to consider during the process~\cite{coston2022validity,jacobs2021measurement}. In particular, recent work has drawn on the concepts such as \emph{validity} and \emph{reliability} from the social science literature~\cite{coston2022validity,jacobs2021measurement} as a way to interrogate choices made during the problem formulation process. As an example, testing for validity ``attempts to establish that a system does what it purports to do''~\cite{coston2022validity}: e.g. as the authors note, establishing validity may be difficult in a predictive policing model that purports to predict \emph{crime}, but in fact predicts new \emph{arrests}, given the large body of work that points to racial disparities in arrest data~\cite{lum2016predict}.

\subsubsection{Case Studies and Problem Identification}
Several recent works have pointed to the impact of problem formulation on equity in model predictions. ~\citet{obermeyer2019dissecting} show that in a health care distribution algorithm meant to identify the sickest patients to recommend them for extra care, the choice to use health care costs as a prediction proxy adds to racial disparities in health care allocation. .~\citet{black2022algorithmic} and ~\citet{benami2021distributive} show that even \textit{how} a given prediction target is formulated---e.g., in the case of Black et al., predicting tax noncompliance to select individuals for audit---as a regression problem (i.e., the dollar amount of misreported tax) or a classification problem (a binary indicator of whether tax noncompliance was over a certain amount) leads to large distributional changes in who is selected by an algorithm, thus impacting algorithmic equity. 
~\citet{benami2021distributive}
also point to the impacts of choosing a prediction universe: the authors show that decisions of what types of permit reports to include in the data can lead to or mitigate disparate impact in environmental remediation due to a concentration of certain types of regulated facilities in areas with higher minority populations. 

\subsubsection{Measurement, Mitigation, and Tools}
Developing protocols for assessing problems of various notions of validity (internal, external, content, predictive) of the target variable, predictive attributes, and prediction universe is a promising and necessary avenue of future work. There is preliminary work along this axis in the form of checklists and protocols, e.g.~\cite{coston2022validity}, but we suggest it may be possible to make a suite of quantitative tests as well.
For example, access to appropriate data, ML practitioners could leverage existing model-level bias-testing frameworks~\cite{AIF360,saleiro2018aequitas} to test for bias across a series of potential prediction tasks to inform the decision of which to choose. Tests for predictive validity simply require investigating whether the proposed target variable is predictive of other related outcomes. In fact, ~\citet{obermeyer2019dissecting} used such a test to uncover the racial bias behind using health care cost as a proxy for health care need, by regressing health care cost on other metrics of illness (e.g. the number of active chronic conditions a patient has), finding that there was a disparity in the correlation between health care costs and sickness in white patients versus that in Black patients. Investigating the extent to which questions of construct validity and reliability can be operationalized into a set of tests (e.g., testing correlations between potential prediction tasks, or checking for consistent model performance across two different methods of measuring the output) may be a promising area for utilizing tools and methods from social sciences.

\subsection{Data Collection}
\subsubsection{Definition and Decisions}
Data collection involves collecting or compiling data to train the model. This involves making choices---or implicitly accepting previously made choices---about how to sample, label, and link data. Some questions include---What population will we sample to build our model? How will we collect this data? What measurement device will be used? How will we link pre-made data?

\subsubsection{Case Studies and Problem Identification}
The algorithmic fairness literature is rife with examples of disparate performance and selection across demographic groups stemming from problems with the training data: for example, datasets that are unbalanced across demographic groups, i.e. have sampling bias, both in terms of sheer representation, and representation conditional on outcomes~\cite{buolamwini2018gender, mishler2019modeling}; have data that is disparately noisy or perturbed in some way~\cite{wang2021fair}; or have labeling bias or untrustworthy labels~\cite{lum2016predict}. 
A string of recent papers test how potential results of data collection problems---e.g. unrepresentative samples, insufficiently small data samples, differing group base rates
---influence machine learning model behavior from the perspective of accuracy and fairness~\cite{li2022data,ding2021retiring,akpinar2022sandbox}. Interestingly, they find that increasing dataset size, or even reducing the disparity in base rates, does not always reduce disparities in selection, false positive, or false negative rates. 

However, fewer papers point to, measure, and show how to mitigate the aspects of the \emph{data collection process itself} that lead to these various forms of biased datasets. ~\citet{paullada2021data} point to failure modes in the data collection process from a high level. The Human and Computer Interaction community has done a more thorough job of considering what failure modes can occur in certain aspects of the data collection process, such as crowdsourcing data set labels with workers from platforms such as Mechanical Turk~\cite{sen2015turkers,hube2019understanding,miceli2020subjectivity}. There are also a few instructive papers that document how sampling and label bias crept into certain real-world ML projects~\cite{marda2020data,sambasivan2021everyone}. For example, ~\citet{marda2020data} show sources of historical bias, sampling bias, measurement bias, and direct discrimination as well as arbitrariness in the collection of data for the creation of Delhi's predictive policing system: for example, for measurement bias, they explain how the techniques used to map Delhi were inconsistent over the development of the tool, since the police department's ArcGIS system license expired.

While this literature does an excellent job of illuminating failure modes that can be surfaced through engaging with ML practitioners and data workers
---since many papers in this area are structured around interviews---they may elide more low-level technical sources of data collection problems, such as record linkage issues leading to non-IID data dropping. While they are helpful in the important first step of identifying problems, they provide less direction for creating operationalizable solutions to these problems, or often even sufficiently detailed information about each system studied to be able to understand the mapping between data collection problems and model behavior. A promising area of future work is to bridge the problems identified by the HCI, CSCW, and other literatures, with the technical detail of the algorithmic fairness literature to introduce mitigation techniques for data collection harms.

\subsubsection{Measurement, Mitigation, and Tools}
There are myriad papers introducing methods of mitigating bias in model predictions given various data problems by manipulating the data or the model. Common methods include data reweighting~\cite{kallus2018residual}, data debiasing~\cite{bolukbasi2016man,kamiran2012data}, synthetic sampling~\cite{sharma2020data, van2021decaf}, and using specialized optimization functions for creating fair classifiers (according to traditional metrics) with unbalanced data~\cite{jiang2021towards, NEURIPS2021_191f8f85}.
There are also less common data interventions, such as one paper by ~\citet{liu2021can} which shows how to find which subgroup in the data is likely to have noisy labels, and then introduce a technique of inserting \emph{more} noise into the labels of other subgroups in order to increase fairness and generalization, and often accuracy as well. 
Another interesting line of work points to how to add additional training samples to the data in order to improve fairness outcomes~\cite{chen2018my, cai2022adaptive}. 

However, there is less work targeting mitigating bias in the data collection process itself. One well-known exception is datasheets for datasets~\cite{gebru2018datasheets}, a paper that introduces a worksheet to fill out when building and distributing a new dataset in order to encourage thought around potential risks and harms as the dataset is being created (e.g. What mechanisms were used to collect the data?), to document potential failure and bias modes for future consumers of the data.
While this work is excellent, to the authors' knowledge, there is little understanding of how well this technique works in practice to prevent data collection mishaps---while preliminary evaluation papers exist~\cite{boyd2021datasheets}, there are none that evaluate its effectiveness in a real-world model building scenario. 
Such evaluation is key to operationalizing these components of pipeline-aware approaches to fairness.

\subsection{Data Preprocessing}
\subsubsection{Definition and Decisions}
Data preprocessing refers to steps taken to make data usable by the ML model--e.g., dropping or imputing missing values, transforming (standardizing or normalizing data), as well as feature engineering, i.e. deciding how to construct features from available information for use in the model (e.g. how to encode categorical, text, image and other data as numbers), and which of the constructed features to use for prediction.

\subsubsection{Case Studies and Problem Identification}
There are myriad ways for biases to enter through data preprocessing decisions. Some of these, particularly those around some aspects of feature engineering, have been explored by the literature---such as creating or selecting features that are differentially informative across demographic populations~\cite{emelianov2022fair,bartik2016deleting, garg2020dropping}, the fairness impacts of including spurious correlations predictive models~\cite{wang2020towards,khani2021removing}, or which choice of features among those available in the data lead to the least disparate impact~\cite{colfax2021report1,colfax2021report2,colfax2022report3}. But other entry points for bias are just beginning to be explored. 
For example, there is little work on understanding the fairness impacts of imputing missing data or dropping rows with missing data. Jeanselme et al.~\cite{jeanselme2022imputation} show that data imputation strategies in the medical context \emph{do} have an impact on accuracy across demographic groups, and that imputation strategies which lead to very similar overall model performance can still lead to different accuracy disparities across groups. Relatedly, Biswas et al.~\cite{biswas2021fair} show, among other effects of preprocessing choices, that dropping rows of a dataset with missing values instead of imputing those rows can lead to sizable differences in fairness behavior due to the changes to the training distribution. These works are an excellent first step, and pave the way for an exciting and necessary avenue for future work: developing methods for how to choose between preprocessing options given information about the data and the modeling pipeline.

Another vastly under-explored area is how data encoding--or how a feature is numerically represented---can have fairness impacts.  Wan~\cite{wan2022fairness} presents an interesting case study of how data representation can affect bias in NLP translation models, and an interesting mitigation technique. They show that differences in translation performance between different languages may not have to do with the structure of language itself, but instead with the granularity of the representation of the language: for example, word length (longer words lead to lower performance).
This performance disparity can be mitigated by using more granular representations of language pieces to equalize representation length across languages (e.g. subwords, different representations for characters). 
While it may be difficult to explore the impacts of different data encoding due to the likely contextualized nature of its effects, we still believe further examples of how such encodings can lead to bias are an important avenue to pursue to further understand sources of bias along the pipeline.

\subsubsection{Measurement, Mitigation, and Tools}
There are several methods of testing for, and mitigating, bias from the inclusion of certain \textit{features}, but tools for isolating and removing bias from other data preprocessing steps are much more understudied.
For example, ~\citet{yeom2018hunting} provide methods for searching for and removing proxies for protected attributes in regression models, and ~\citet{frye2020asymmetric} introduce Asymmetric Shapley Values, a technique that can be used on a wider range of models to determine whether a feature is allowing a protected attribute to causally influence the model outcome, so that the feature can then be pruned. 
Additionally, the FlipTest technique~\cite{fliptest} also produces suggestions of which features may be the source of disparate outcomes across any two populations the user may wish to compare, without attention to causality---this may be especially helpful in narrowing down the list of features to further investigate for impacts on differences in selection rate, or simply to find statistical, and not causal, discrimination.  Finally, ~\citet{belitz2021automating} provide an automatic feature selection framework that only selects features that improve accuracy without reducing a user-specified notion of fairness. However, this method saw considerable drops in accuracy when selecting features in this manner. We note that even among feature-related measurement and mitigation tools, there is little work surrounding how to identify and mitigate bias from the use of features with different variances or predictive power across demographic groups.

Despite the capacity for feature engineering and imputation choices to affect model fairness, our survey failed to identify any tools that assist practitioners in mapping out the effects of their preprocessing choices from the perspective of preventing bias and suggesting mitigations. However, there may be some potential for adapting tools that have been developed for more general data exploration and preprocessing to incorporate fairness contributions as well. For instance, ~\citet{breck2019data} describe a data validation pipeline for ML used at Google that proactively searches for data problems such as outliers and inconsistencies; and there are several data-cleaning tools that even suggest transformations according to best practices~\cite{hynes2017data, krishnan2016activeclean}. Further exploring the landscape of these more general tools and their potential applicability to questions of model fairness seems to be a fruitful avenue for future work. Crucially, however, we note that we are unaware of any such systems that presently account for bias-related desiderata.

\subsection{Statistical Modeling}
\subsubsection{Definition and Decisions}
After deciding how to preprocess data, model makers must decide how they will create a model for their data and how it will be trained.  Decisions here include what type of model will be used, the learning rule and loss function, regularizers, the values of hyper-parameters determining normalization and training procedures, among other decisions made continuously throughout model development~\cite{lehr2017playing, kim2022race}. For example, choosing between linear, forest-based, or deep models; singular models or ensembles, choosing the architecture of deep models; what constraints to add to the loss function, how to optimize that loss function (e.g. SGD or momentum), among many other choices. 

\subsubsection{Case Studies and Problem Identification}
The majority of algorithmic fairness papers show how to \emph{intervene} on a model's loss function or prediction process to reduce biased behavior--- but few papers point to sources of bias \textit{stemming} from a model's loss function and other modeling decisions, and how to identify such problems. However, \emph{every} such decision can lead to downstream bias: for example, a chosen learning rule might over-rely on certain features, leading to skewed predictions for certain populations~\cite{leino2018featurewise}, or may over- or under-emphasize outliers or minority populations. Model type selection has been shown to impact fairness (e.g., high-capacity models with increased variance can be more unstable to small perturbations in training setup~\cite{blackleave2021} leading to procedural fairness concerns about the nature of the decision process). A growing number of works have pointed to degradations of fairness behavior in robust models~\cite{xu2021robust,ma2022tradeoff,10.1145/3514094.3539552} and differentially private models~\cite{bagdasaryan2019differential, suriyakumar2021chasing}.
~\citet{d2020underspecification} point to the importance of loss function choice in fairness behavior: they show without explicitly specifying a desired behavior--including fairness--within a model's loss function, the resulting model is unlikely to naturally display near-optimal or even acceptable behavior on that desired property.

\subsubsection{Measurement, Mitigation, and Tools}
Most of the focus on fairness interventions in statistical modeling is centered around changing model loss function or prediction process to enforce fairness constraints~\cite{agarwal2018reductions, AIF360}. However,  more recently, these conventional techniques which enforce group and individual fairness metrics on top of a decision system have garnered criticism~\cite{akpinar2022sandbox,suresh2021framework,black2022algorithmic}, and there has been evidence showing that enforcing such constraints can actually degrade fairness according to those same metrics~\cite{wang2021fair}. Outside of intervening on the loss function, there are few papers that introduce mitigation techniques for bias introduced at other stages of the statistical modeling process. Notable exceptions include ~\citet{islam2021can} and ~\citet{perrone2021fair}, who show that hyperparameter tuning can lead to increased fairness at no cost to accuracy. ~\citet{perrone2021fair} introduce a technique, Fair Bayesian Optimization (FBO), which is model and fairness-definition agnostic, to select hyperparameters that optimize accuracy subject to the fairness constraint. They also experimentally demonstrate that regularization parameters are the most influential for fairness performance, and that higher regularization leads to higher fairness performance in their experiments. We hope that this technique can be used to understand further relationships between hyperparameter changes and fairness behaviors over a variety of different models, metrics, and deployment scenarios. However, as has been a common pattern, we did not discover papers that developed tools to \emph{measure} the extent to which statistical modeling choices impacted model fairness behavior. 

\subsection{Testing and Validation}
\subsubsection{Definition and Decisions}
Model testing and validation refers to the processes by which a model is determined to be performing well, both in relation to other models in the training set, but also on unseen data. Some decisions here include whether a model be evaluated only on its predictive accuracy, AUC, F1 score, or another performance metric; on fairness metrics as well (and choosing which); or on some notion of privacy or robustness. It also includes decisions such as what size of the dataset will be reserved for evaluation (train/test/evaluation split); what datasets the model will be evaluated on; and how many trials or k-folds the model will be evaluated on.

\subsubsection{Case Studies and Problem Identification}
Several papers have discussed the perils of evaluating systems on the same benchmark data sets: this can lead to overfitting to specific data sets~\cite{paullada2021data} that almost guarantees distribution shift to deployment domains, meaning that results are unrepresentative for many real-world fairness applications~\cite{sambasivan2021everyone}, and suggests that several experimental results in the fairness literature may be incorrect~\cite{ding2021retiring,li2022data}. 
 

Others have suggested issues with the metrics used to compute bias in ML systems themselves--e.g. Lum et al.~\cite{lum2022biasing} show that many such metrics ``are themselves statistically biased estimators of the underlying quantities they purport to represent''; others have shown that under circumstances such as label bias or extreme feature noise, enforcing fairness metrics can actually lead to \emph{decreased} fairness behavior along those very same metrics in the model~\cite{wang2021fair}.  We believe there may be other sources of unfairness in the testing and validation part of the pipeline---such as mismatching testing and validation metrics to the application context (as discussed in ~\cite{black2022algorithmic}); but we were unable to find studies of any such problems on the ground. 
\subsubsection{Measurement, Mitigation, and Tools}
Several frameworks allow ML practitioners to test for bias in their models' predictions~\cite{saleiro2018aequitas,AIF360,hardt2021amazon,10.1145/3351095.3375662}, however we note that these frameworks do not target which part of the pipeline may be adding to this bias, it only allows for bias testing to occur along the most popular fairness metrics. These frameworks allow for the most basic bias mitigation to testing and validation problems: not testing for fairness during validation at all. Several recent works have also added new or expanded datasets to be used as benchmarks in fairness contexts to prevent problems of dataset overfitting~\cite{ding2021retiring,li2022data}, though fairness researchers may benefit from exploring datasets even beyond these, perhaps collaborating or borrowing data from social scientists, or exploring many of the less popular publicly available dataset such as [environmental dataset],[american community survey]. However, there were no papers that we could find during our survey which showed how to measure the extent to which testing and validation design choices influence downstream fairness behavior.

\subsection{Deployment and Monitoring}

\subsubsection{Definition and Decisions}
Deployment refers to the process of embedding a model into a larger decision system. 
For example, some decisions here include: how will the model be used as a component of the decision system into which it is embedded? Will the model's predictions directly become the final decision? If there is human involvement, where and how will that occur? How much discretion do humans have over adhering to model recommendations? How are model predictions communicated to decision-makers? 
Monitoring refers to how a model's behavior is recorded and responded to over time in order to ensure there is no degradation in performance over time, fairness or otherwise. Decisions here include: Will monitoring occur? If so, how will performance over time be measured? How and when will data drift be measured and addressed?

\subsubsection{Case Studies and Problem Identification}
There has been a stream of theoretical work from the algorithmic fairness literature trying to model or guarantee fairness in a joint human-ML system~\cite{keswani2021towards,madras2018predict}. For example, ~\citet{keswani2021towards} learn a classifier and a deferral system for low-confidence outputs, with the deferral system taking into account the biases of the humans in the loop.~\citet{donahue2022human} develop a theoretical framework for understanding when and when not human and machine error will complement each other. While this initial largely theoretical work is promising, these techniques should be tested and compared on deployed systems---when do deferral systems work in practice, and can they lead to biases of their own? Do the human-in-the-loop design suggestions work in practice?

However, there are conflicting results as to whether human-in-the-loop systems outperform models on their own when it comes to bias. ~\citet{green2019principles,green2019disparate} show that including Mechanical Turk workers to aid algorithmic decisions consistently decreased accuracy relative to the algorithm's performance alone, and that humans in the loop also exhibited racial bias when interacting with ML predictions. However, others~\cite{cheng2022child, de2020case} have found that in the case of child welfare screenings, allowing call screen workers \emph{did} reduce disparity in the screen-in rates of Black versus white children. 

\subsubsection{Measurement, Mitigation, and Tools}
While a few papers provide mitigation techniques and tools for fairness monitoring and preventing distribution shifts, we note that no papers that appeared in our survey provided techniques for addressing biases that arise as a result of including humans in ML decision systems---an important area of future work. 

A series of recent papers have introduced methods to make models invariant to distribution shift from the perspective of accuracy as well as fairness~\cite{singh2021fairness,coston2019fair,biswas2021ensuring}. However, there are several open questions in this area of research: such as, when do we want to ensure fairness criteria over data drift, and when do we want to alert the model practitioner that the distributional differences are so large that the model is not suitable for the deployment context?

Pertinently, two recent papers~\cite{albarghouthi2019fairness,ghosh2022faircanary} propose tools for fairness monitoring over time in deployment. ~\citet{albarghouthi2019fairness} develop a technique called fairness-aware programming 
which allows programmers to enforce probabilistic statements over the behaviors of their functions, and get notified for violations of those statements---their framework is flexible to many behavioral desiderata even beyond fairness, and can also combine requirements and check 
them simultaneously. The flexibility of the system potentially allows for a wide range of contextualized fairness desiderata to be enforced-- however, it does have limitations; for example, it cannot implement notions of fairness over individual inputs such as individual fairness. 
~\citet{ghosh2022faircanary} implement a system that measures the Quantile Demographic Drift metric at run time, a notion of fairness that can shift between group and individual conceptions of fairness based on the granularity of the bins it calculates discrepancies over. Their system also offers automatic mitigation strategies (normalizing model outputs across demographic groups) and explanation techniques to understand mechanisms of bias. Additionally, Amazon's SageMaker Clarify framework allows for the tracking of a variety of traditional fairness metrics to be monitored at runtime~\cite{hardt2021amazon}. 



\section{Discussion and A Proposal for Pipeline Cards}
\label{sec:agenda}
Leveraging our survey of the literature, we move towards creating operational pipeline-aware fairness techniques by 
developing a research agenda towards operationalizing pipeline-aware fairness techniques, and introducing \emph{Pipeline Cards}, a documentation system for interrogating design choices in the ML pipeline, in the style of previous fairness work~\cite{mitchell2019model,gebru2018datasheets}. While we leave many details to Appendices 2 and 3, we summarize our contributions here.

\xhdr{Research Agenda} Reflecting on the state of fairness literature from a pipeline-aware perspective, we offer five key insights to start operationalizing a pipeline-aware approach to fairness, which we expand into a larger research agenda in Appendix 2. Namely, we identify (1) the need to investigate and document choices made along real-world pipelines, including those related to bias mitigation; (2) the need to bridge the gap across literatures which \emph{identify} ways the bias enters ML pipelines on-the-ground, such as HCI, and literatures which build operational mitigation techniques, such as FairML; (3) the need to study interaction effects across decisions made along the pipeline; (4) the need to address holes in the current research--paying attention to neglected areas of the pipeline such as viability assessment, and entire modes of research such as producing \emph{measurement} techniques to catch many of the entry points for bias identified along the pipeline; and finally, (5) the need to produce guidance on how to choose among several biased building choices.

\xhdr{Pipeline Cards}
To provide an actionable step towards our research agenda, we provide a first version of a tool that ML practitioners can use to build AI systems with a pipeline-aware approach to fairness. In the spirit of Model Cards and Data Sheets~\cite{gebru2018datasheets,mitchell2019model} we introduce \emph{Pipeline Cards}: a documentation framework for design choices along the ML pipeline to promote practitioner introspection and transparent reporting~\cite{black2022model} of how AI systems are made. Pipeline Cards can be used in conjunction with our Pipeline Fairness wiki\footnote{http://fairpipe.dssg.io/} to think through potential fairness problems in design choices, and then search for relevant measurement and mitigation methods. We present our current version of Pipeline Cards in Appendix 3. 
Pipeline Cards are under active development, including testing with practitioners and students to understand how best to prompt practitioners to isolate potential problems in the ML pipeline, which we look forward to presenting in future work.


\bibliography{bib}
\bibliographystyle{ACM-Reference-Format}
\newpage 


\begin{table}[]
\centering
\caption*{Pipeline-Aware Taxonomy of Literature Survey} 
\resizebox{\textwidth}{!}{%
\begin{tabular}{|l|l|l|l|l|l}
\hline

 Stage &
  Step &
  Problem Identification &
  Measurement &
  Mitigation \\ \hline\hline
Viability Assessments &
  Cost/Benefit &
   &
   &
   \\
 &
  General & \cite{raji2022fallacy}, \cite{barocas2020not}, \cite{passi2020making}
   & \cite{wang2022against}, \cite{coston2022validity}
   &
   \\ \hline
Problem Formulation &
  Prediction Target &
  \begin{tabular}[c]{@{}l@{}} \colorbox{magenta}{\cite{black2022algorithmic}},\colorbox{magenta}{\cite{obermeyer2019dissecting}}, \colorbox{magenta}{\cite{human_rights_watch_2020}},\\ \colorbox{magenta}{\cite{10.1145/3531146.3534644}} \end{tabular} &
   & \cite{benami2021distributive}
   \\ \hline
 &
  Predictive Attributes & \cite{grgic2018beyond}
   &
   &
   \\ \hline
 &
  General & 
  \begin{tabular}[c]{@{}l@{}}\colorbox{magenta}{\cite{robertsonchi21}},  \colorbox{magenta}{\cite{zhangchi22}}, \colorbox{magenta}{\cite{navarrocscw22}}, \colorbox{magenta}{\cite{nikkhahcscw22}}, \cite{passi2019problem} \\, \colorbox{magenta}{\cite{pmlr-v81-chouldechova18a}},  \cite{10.1145/3442188.3445901}, \cite{chengchi21}, \colorbox{magenta}{\cite{ilventofat20}}, \cite{lum2016predict}
  \end{tabular}
   &
   &
   \\ \hline
Data Collection &
  Sampling &
  \begin{tabular}[c]{@{}l@{}} \colorbox{magenta}{\cite{buolamwini2018gender}}, \colorbox{magenta}{\cite{10.1145/3442188.3445916}}, \colorbox{magenta}{\cite{10.1145/3531146.3533132}}, \colorbox{magenta}{\cite{10.1145/3531146.3533159}}, \\ , \cite{cai2022adaptive} \end{tabular} & 
  \cite{mishler2019modeling}, \cite{yik2022identifying} &
  \begin{tabular}[c]{@{}l@{}}~\cite{sharma2020data}, \cite{van2021decaf}, \cite{10.1145/3442188.3445940}, \colorbox{yellow}{\cite{tsai2022conditional}}, \colorbox{yellow}{\cite{shekharneurips21}}, \colorbox{yellow}{\cite{roh2021fairbatch}} \end{tabular} \\ \hline
 &
  Annotation &
  \begin{tabular}[c]{@{}l@{}}\colorbox{magenta}{\cite{wei2021learning}}, \cite{wang2021fair},
  \cite{10.1145/3531146.3533216}, \cite{christoforouchi21}\end{tabular} &
  ~\cite{liu2021can}, \cite{10.1145/3531146.3533099}, \cite{birhane2021values} &
  \cite{10.1145/3351095.3375709} \\ \hline
 &
  Feature Measurement &
  \begin{tabular}[c]{@{}l@{}} \colorbox{magenta}{\cite{wan2022fairness}}, \colorbox{magenta}{\cite{10.1145/3461702.3462561}}, \cite{frye2020asymmetric} \end{tabular}
   &
  \cite{10.1145/3306618.3314270} &
   \\ \hline
 &
  Record Linkage &
   &
   &
   \\ \hline
 &
  General &
  \begin{tabular}[c]{@{}l@{}} \colorbox{magenta}{\cite{NEURIPS2021_e97a4f04}}, \colorbox{magenta}{\cite{hettiachchicscw21}},
  \colorbox{magenta}{\cite{sambasivan2021everyone}}, \cite{10.1145/3442188.3445870}, \\ 
  \cite{kallus2018residual}, \cite{marda2020data} \end{tabular} &
  \begin{tabular}[c]{@{}l@{}}\cite{li2022data}, \cite{ding2021retiring}, \cite{NEURIPS2021_85dca1d2}, \cite{10.1145/3461702.3462590}, \cite{NEURIPS2020_d83de59e},\\ \cite{10.1145/3351095.3373154}, \cite{10.1145/3442188.3445877}, \cite{bolukbasi2016man}, \cite{kamiran2012data} \end{tabular} &
  ~\cite{chen2018my}, \cite{gebru2018datasheets}, \cite{boyd2021datasheets} \\ \hline
Data Preprocessing &
  Feature Creation &
   &
   &
   \\ \hline
 &
  Feature Selection &
  \cite{10.1145/3461702.3462585}, \cite{NEURIPS2018_6cd9313e}, \cite{fliptest}, \cite{leino2018featurewise}   &
  \cite{10.1145/3461702.3462585}, \cite{NEURIPS2019_201d5469} &
  \cite{NEURIPS2018_6cd9313e}, \cite{10.1145/3531146.3533126}, \colorbox{yellow}{\cite{heaies20}} \\ \hline
 &
  Data Cleaning (Omission) &
   &
  \cite{NIPS2017_8cb22bdd}, \cite{breck2019data} &
  \cite{NIPS2017_9a49a25d}, \cite{10.1145/3442188.3445878}, \cite{hynes2017data}, \cite{krishnan2016activeclean}  \\ \hline
 &
  General &
  \begin{tabular}[c]{@{}l@{}}\cite{10.1145/3514094.3534162}, \cite{NEURIPS2021_d800149d}, \cite{pmlr-v139-liang21a}, \cite{biswas2021fair}\end{tabular} &
  \begin{tabular}[c]{@{}l@{}}\cite{10.1145/3531146.3533243}, \cite{10.1145/3531146.3534632}, \cite{pmlr-v119-choi20a}, \cite{pmlr-v162-shui22a}, \\ \cite{pmlr-v162-wang22ac}, \cite{pmlr-v162-ganev22a} \end{tabular} &
  \begin{tabular}[c]{@{}l@{}}\cite{NIPS2017_e6384711}, \cite{NEURIPS2020_af9c0e0c},  \cite{NEURIPS2021_191f8f85}, \cite{NEURIPS2021_64ff7983},\\ \cite{10.1145/3351095.3372837}, \cite{10.1145/3351095.3372843}, \cite{10.1145/3531146.3533175}, \cite{pmlr-v97-creager19a},\\ \cite{pmlr-v119-buyl20a}, \cite{pmlr-v119-celis20a}, \cite{pmlr-v139-trauble21a}, \cite{pmlr-v162-kim22b},  \end{tabular} \\ \hline
Statistical Modeling &
  Hypothesis Class &
  \cite{10.1145/3442188.3445894}, \cite{10.1145/3442188.3445910},   &
  \cite{zhu2022the}  &
  \cite{10.1145/3514094.3534142} \\ \hline
 &
  Optimization Function &
  \begin{tabular}[c]{@{}l@{}}\cite{10.1145/3351095.3372872}, \cite{suriyakumar2021chasing}, \cite{pmlr-v80-hashimoto18a}, \cite{pmlr-v162-dasgupta22b}, \\
  \end{tabular} 
  & \cite{d2020underspecification}
  & \begin{tabular}[c]{@{}l@{}} \cite{xu2021robust} , \colorbox{yellow}{\cite{russellneurips17}}, \colorbox{yellow}{\cite{aliaies19}}, \\ \colorbox{yellow}{\cite{currenteaamo22}}, 
  \colorbox{yellow}{\cite{roh2021fairbatch}}, \colorbox{yellow}{\cite{romanoneurips20}}, \\ \colorbox{yellow}{\cite{sharifmalvajerdineurips19}}, \colorbox{yellow}{\cite{chzhenneurips20}}, \colorbox{yellow}{\cite{choneurips20}}, \colorbox{yellow}{\cite{rohneurips21}}, \\ \colorbox{yellow}{\cite{louppeneurips17}}, \cite{li2021tilted}, \cite{xuneurips21},  
  \end{tabular}\\ \hline
 &
  Regularizers &
   &
   &
  \cite{pmlr-v162-kancheti22a}, \cite{pmlr-v162-shah22a} \\ \hline
 &
  Hyperparameters &
  \cite{10.1145/3375627.3375807}  &
   &
  \begin{tabular}[c]{@{}l@{}}\cite{NEURIPS2021_18de4beb}, \\ \cite{10.1145/3531146.3533094},\cite{Schmucker2020}, \cite{islam2021can},\cite{perrone2021fair} \end{tabular} \\ \hline
 &
  General &
  \begin{tabular}[c]{@{}l@{}}\cite{NEURIPS2021_08425b88}, \cite{10.1145/3531146.3534641}, \cite{pmlr-v119-lohaus20a},\\ \cite{pmlr-v139-coston21a}, \cite{beutelaies19} \\ 
  
  \cite{akpinar2022sandbox}, \cite{bagdasaryan2019differential}, \colorbox{magenta}{\cite{10.1145/3514094.3539552}}
  \end{tabular} 
  &
  \begin{tabular}[c]{@{}l@{}} \cite{navon2021learning}, \cite{10.1145/3287560.3287594}, \cite{10.1145/3351095.3372853},\\ \cite{pmlr-v139-wang21t} , \cite{ma2022tradeoff}
  \end{tabular} &
  \begin{tabular}[c]{@{}l@{}} \cite{NEURIPS2021_4b26dc46}, \cite{NEURIPS2019_8d5e957f}, \cite{NEURIPS2020_55d491cf}, \\  \cite{li2021on}, \cite{balunovic2022fair}, \cite{pmlr-v81-dwork18a},\\ \cite{pmlr-v81-kamishima18a}, \cite{pmlr-v81-burke18a}, \cite{10.1145/3442188.3445887},\cite{10.1145/3531146.3533136},\\ \cite{pmlr-v97-nabi19a}, \cite{pmlr-v162-foster22a},  \\ \colorbox{yellow}{\cite{zhangaies18}}, \colorbox{yellow}{\cite{kimaies19}}, \colorbox{yellow}{\cite{sharmaaies21}}, \colorbox{yellow}{\cite{halevyeaamo21}} \\ \colorbox{yellow}{\cite{lahotineurips20}},  
  \colorbox{yellow}{\cite{petersenneurips21}}, \colorbox{yellow}
  {\cite{Zhao2020Conditional}}, \\
  \colorbox{yellow}{\cite{salvador2022faircal}}, \colorbox{yellow}{\cite{wangfacct22}},
  \colorbox{yellow}{\cite{mishlerfadefacct22}},
  \colorbox{yellow}{\cite{wufacct22}} \\
  \colorbox{yellow}{\cite{agarwal2018reductions}}, \colorbox{yellow}{\cite{pmlr-v80-kilbertus18a}}, \colorbox{yellow}{\cite{pmlr-v80-madras18a}}, \colorbox{yellow}{\cite{chuang2021fair}}, \\ \colorbox{yellow}{\cite{celis2019classification}}, \colorbox{yellow}{\cite{gargaies19}}, \colorbox{yellow}{\cite{pmlr-v80-komiyama18a}},
  \colorbox{yellow}{\cite{onetoaies19}}, \\ \cite{aliaies21}, \colorbox{yellow}{\cite{grabowiczfacct22}}, \colorbox{yellow}{\cite{fisheaamo22}}, \colorbox{yellow}{\cite{chenaies20}}, \colorbox{yellow}{\cite{thomaseaamo21}}, \colorbox{yellow}{\cite{quadriantoneurips17}}, \colorbox{yellow}{\cite{kusnerneurips17}} \\ \colorbox{yellow}{\cite{doninineurips18}}, \colorbox{yellow}{\cite{heidarineurips18}}, \colorbox{yellow}{\cite{bower2021individually}},\colorbox{yellow}{\cite{vargo2021individually}}, \\ \colorbox{yellow}{\cite{yurochkin2021sensei}}, \colorbox{yellow}{\cite{xu2022controlling}}, \colorbox{yellow}{\cite{NEURIPS2019_d69768b3}}, 
  \colorbox{yellow}{\cite{narasimhanneurips20}}, \\
  \colorbox{yellow}{\cite{caycineurips20}}, 
  \colorbox{yellow}{\cite{10.1145/3442188.3445865}}, \colorbox{yellow}{\cite{elzaynfat19}}, \colorbox{yellow}{\cite{pmlr-v80-kearns18a}}, \\ \colorbox{yellow}{\cite{pmlr-v80-yona18a}}, \colorbox{yellow}{\cite{pmlr-v97-cotter19b}}, \colorbox{yellow}{\cite{goelaies18}}, \colorbox{yellow}{\cite{raffaies18}}, \\ \colorbox{yellow}{\cite{liuaies21}}, \colorbox{yellow}{\cite{shahguptaaies21}}, \colorbox{yellow}{\cite{liuneurips19}}, \colorbox{yellow}{\cite{chzhendenisneurips20}}, \\ \colorbox{yellow}{\cite{bechavodneurips20}},
  \colorbox{yellow}
  {\cite{doneurips21}},
  \colorbox{yellow}{\cite{mishlerfacct21}}, \colorbox{yellow}{\cite{usunierfacct22}}, \\ 
  \colorbox{yellow}{\cite{pmlr-v80-hebert-johnson18a}},
  \colorbox{yellow}{\cite{pmlr-v97-agarwal19d}}, \cite{NEURIPS2019_373e4c5d}, \colorbox{yellow}{\cite{farnadiaies18}},  \colorbox{yellow}{\cite{harpeledneurips19}},  \colorbox{yellow}{\cite{Baharlouei2020Rényi}}, \colorbox{yellow}{\cite{Yurochkin2020Training}}, \cite{NEURIPS2021_dcf531ed},
  \end{tabular} \\ \hline 
Testing and Validation &
  Train-test split &
   &
   &
   \\ \hline
 &
  Evaluation Metrics &
  \begin{tabular}[c]{@{}l@{}}\cite{NEURIPS2020_7ec2442a}, \cite{dullerud2022is}, \cite{10.1145/3531146.3533149},   \colorbox{yellow}{\cite{zhangmarilynfacct22}}, \cite{NEURIPS2021_ed277964},
  \\  \end{tabular} & 
  \begin{tabular}[c]{@{}l@{}} \colorbox{yellow}{\cite{michalskyaies19}}, \colorbox{yellow}{\cite{sharmaaies20}}, \colorbox{yellow}{\cite{dianaaies21}}, \colorbox{yellow}{\cite{kimneurips18}}, \\ \colorbox{yellow}{\cite{zhangneurips18}}, \colorbox{yellow}{\cite{kallusneurips19}}, \colorbox{yellow}{\cite{smithneurips20}}, 
  \colorbox{yellow}{\cite{maity2021statistical}} \\ 
  \colorbox{yellow}{\cite{jiang2022generalized}}, 
  \colorbox{yellow}{\cite{heringtonaies2020}}, 
  \colorbox{yellow}{\cite{cousinsneurips21}},
  \colorbox{yellow}{\cite{costonfat20}}, \colorbox{yellow}{\cite{gargaies19}}, \colorbox{yellow}{\cite{mutluaies22}}, \cite{10.1145/3442188.3445884},   \cite{10.1145/3442188.3445927}, \cite{10.1145/3531146.3533102}, \cite{pmlr-v162-yan22c},
  \end{tabular}
   &
   \begin{tabular}[c]{@{}l@{}} \cite{NEURIPS2018_415185ea}, \colorbox{yellow}{\cite{mhasawadeaies21}}, \colorbox{yellow}{\cite{zafarneurips17}}, \colorbox{yellow}{\cite{kilbertusneurips17}}, \colorbox{yellow}{\cite{canettifat19}}, \colorbox{yellow}{\cite{chzhenneurips19}}, \\ \colorbox{yellow}{\cite{balcanneurips19}}, \colorbox{yellow}{\cite{yangneurips20}}, \colorbox{yellow}{\cite{chenneurips20}}, \colorbox{yellow}{\cite{yanglorchneurips20}}, \\ \colorbox{yellow}{\cite{huangneurips21}}, \colorbox{yellow}{\cite{bendekgeyneurips21}}\end{tabular} \\ \hline
 & 
  General &
  \begin{tabular}[c]{@{}l@{}} \colorbox{magenta}{\cite{pmlr-v80-olofsson18a}}, \cite{pmlr-v139-si21a} \end{tabular} &
  \cite{10.1145/3351095.3372845}, \cite{lum2022biasing}, \cite{AIF360}, \cite{hardt2021amazon}, \cite{paullada2021data}, \cite{saleiro2018aequitas}, \colorbox{yellow}{\cite{kimfat20}}  &
  \begin{tabular}[c]{@{}l@{}} \cite{10.1145/3306618.3314236}, \cite{10.1145/3278721.3278729},\\ \cite{cheng2021fairfil}, \cite{pmlr-v97-wang19l},  \end{tabular} \\ \hline 
Deployment and Integration &
  Human-Computer Handoff &
  \begin{tabular}[c]{@{}l@{}} \colorbox{magenta}{\cite{jacobs2021machine}}, \cite{keswani2021towards}, \cite{green2019disparate}, \cite{10.1145/3531146.3533152}, \cite{10.1145/3531146.3533221}, \\ \cite{leungchi2020}, \cite{machi22}, \cite{watkinscscw20} \end{tabular} & \cite{cheng2022child}
   & \cite{anikchi21}
   \\ \hline
 &
  Maintenance &
   & 
   & 
   \\
 &
  Oversight &
   &
  \cite{albarghouthi2019fairness}, &
   \\ \hline
 &
  General & \begin{tabular}[c]{@{}l@{}} \colorbox{magenta}{\cite{10.1145/3351095.3372827}}, \colorbox{magenta}{\cite{10.1145/3351095.3372846}},
  \cite{10.1145/3514094.3534173}, \cite{parkchi22}, \\ 
  \cite{madaio2020co}, \cite{green2019principles}, \cite{green2019disparate}, \cite{cheng2022child}, \cite{de2020case} \end{tabular} &
  \begin{tabular}[c]{@{}l@{}} \cite{ghosh2022faircanary}, \cite{giguere2022fairness},\cite{10.1145/3351095.3372839}, \cite{10.1145/3531146.3533230},\\ \cite{pmlr-v119-hinder20a},  \cite{biswas2021ensuring} \end{tabular} &
  \begin{tabular}[c]{@{}l@{}}\cite{10.1145/3351095.3375783}, \cite{10.1145/3442188.3445865}, \cite{10.1145/3531146.3533211}, \colorbox{yellow}{\cite{camperoaies19}}, \cite{suresh2021framework}, \cite{madras2018predict}, \cite{coston2019fair} \end{tabular} \\ \hline
\end{tabular}%
}
\caption{A taxonomy of the papers surveyed into the various sections of the ML pipeline they study, and whether they identify, measure, or mitigate a source of bias. The pink papers correspond to case studies within problem identification. Yellow colored references denote papers that correspond to more traditional approaches to fairness, i.e. imposing a fairness constraint on top of a pre-made modeling process, or introducing a new notion of fairness in the testing and evaluation section.}
\label{tab:survey-table}
\end{table}

\section{Classroom Study}
\label{app:classroom}
Here, we present further details on the population and sampling and study design of the classroom study in Section~\ref{sec:teaching_evidence}. 

\xhdr{Population and sampling. }
In total, 37 students participated in the class activity. 
 Prior completion of at least one Machine Learning course was a prerequisite for enrollment. So all participants had a non-negligible background in Machine Learning. Specifically, $\sim 80\%$ characterized the familiarity with ML as intermediate, $\sim 14\%$ as advances, and $\sim 5\%$ as elementary. All students enrolled in the course took it as an elective. This fact implies that compared to a random sample of students with ML knowledge, our participants were likely more aware of ML harms and more motivated to address them. 

\xhdr{Study design.} Students were introduced to the ML pipeline through an approximately 45-minutes long lecture. The lecture focused on the supervised learning paradigm and broke down the supervised learning pipeline into the following five stages: 

\begin{enumerate}
    \item \textbf{Problem Formulation} corresponding to the choice of features and predictive targets.
    \item \textbf{Data collection and processing} corresponding to the choice and processing of the input data $D = \{(x_i,y_i)\}_{i=1}^n$
    \item \textbf{Model specification} corresponding to the choice of the hypothesis class, $H$.
    \item \textbf{Model fitting/training:} choosing a specific loss function, $L$, and optimizing it via existing optimization techniques to obtain a model $h \in H$
    \item \textbf{Deployment in real-world}, corresponding to the deployment of $h$ in real-world practice, and its predictions $\hat{y}$ translating into decisions that can harm/benefit certain individuals and communities
\end{enumerate}

For each stage, students were presented with multiple examples of how choices at that stage can lead to harmful outcomes, such as unfairness, at the end of the pipeline (see Figure~\ref{fig:pipeline}). 
They were then asked to team up with 3-4 classmates and pick a societal domain as the focus of their group activity. Examples offered to them included employment (e.g., hiring employees); 
education (e.g., admitting students); 
medicine (e.g., diagnosing skin cancer); 
housing (e.g., allocating limited housing units
child welfare (e.g., investigating referral calls); 
criminal justice (e.g., pretrial sentencing);
public safety (e.g., allocating patrol resources in policing 
e-commerce (e.g., advertising; ranking sellers on Amazon); 
social media (e.g., news recommendation; content moderation);
and transportation (e.g., autonomous vehicles). One team suggested finance (e.g., credit lending) as their topic.
Third, students were given 30 minutes to discuss the following questions about their application domain with teammates and submit their written responses individually:
\begin{itemize}
    \item Characterizing the specific \textbf{predictive task} their team focused on. 
    \item The \textbf{type of harm} observed. 
    \item Their \textbf{hypotheses around the sources} of this harm in through the ML pipeline. 
    \item Their\textbf{hypotheses around potentially effective remedies} for addressing those sources. 
\end{itemize}

We present an overview of the application domains that students discussed, and how they defined the predictive task in their area. Our findings are discussed in Section~\ref{sec:teaching_evidence}.


\begin{table}[]
{\footnotesize

\caption{An overview of the predictive tasks students chose to analyze through a pipeline-centric view of the ML pipeline.}
\label{tab:classroom}
\begin{tabular}{|l|c|l|l}
\cline{1-3}
\textbf{Domain} & \multicolumn{1}{l|}{\textbf{Students}} & \textbf{Predictive Task}                                                                                                                                                                                                                                                                      &  \\ \cline{1-3}
Child Welfare   & 3                                      & \begin{tabular}[c]{@{}l@{}}Predicting the risk of a child running away from their foster care within 90 days of being in the child welfare system, \\ based on a combination of demographic and clinical characteristics, and information about them in the welfare system.\end{tabular}      &                          \\ \cline{1-3}
Education       & 4                                      & \begin{tabular}[c]{@{}l@{}}Predicting whether an individual is cheating in an online exam \\ based on visual and auditory cues (e.g., irregular eye/body movement, mouse movement, facial expression, and noise).\end{tabular}                                                                &                          \\ \cline{1-3}
Employment      & 6                                      & Predicting whether a company will hire an applicant based on the information in their resume.                                                                                                                                                                                                 &                          \\ \cline{1-3}
Finance         & 3                                      &  Assessing the credit worthiness of individuals using their demographic \&  socio-economic data.                                                                                                                                                                        &                          \\ \cline{1-3}
Housing         & 3                                      & Predicting the value of a real-estate properties                                                                                                                                                                                                                                              &                          \\ \cline{1-3}
Medicine        & 8                                      & \begin{tabular}[c]{@{}l@{}}1) Predicting which people will/should receive an organ transplant.\\ 2) Predicting prognosis (e.g., mortality) for comatose patients post-cardiac arrest \\ using demographic, medical history, and medical screening data (e.g., CT scan, EEG data)\end{tabular} &                          \\ \cline{1-3}
Public Safety   & 4                                      & \begin{tabular}[c]{@{}l@{}}1) Allocating police presence to high risk areas; \\ 2) Identifying suspicious individuals given a list of wanted criminals.\end{tabular}                                                                                                                          &                          \\ \cline{1-3}
Social Media    & 4                                      & Predicting whether a news is fake.                                                                                                                                                                                                                                                            &                          \\ \cline{1-3}
Transportation  & 2                                      & Detecting objects (e.g., human, vehicles, road conditions) for AVs.                                                                                                                                                                                                                           &                          \\ \cline{1-3}
\end{tabular}
}
\end{table}

\section{Research Agenda}
\label{app:agenda}


Reflecting on the state of fairness literature from a pipeline-aware perspective, we offer five key insights to start operationalizing a pipeline-aware approach to fairness. Namely, we identify (1) the need to investigate and document choices made along real-world pipelines, including those related to bias mitigation; (2) the need to bridge the gap across literatures which \emph{identify} ways the bias enters ML pipeliens on-the-ground, such as HCI, and literatures which build operational mitigation techniques, such as FairML; (3) the need to study interaction effects across decisions made along the pipeline; (4) the need to address holes in the current research--paying attention to neglected areas of the pipeline such as viability assessment, and entire modes of research such as producing \emph{measurement} techniques to catch many of the entry points for bias identified along the pipeline; and (5) finally, the need to produce guidance on how to choose among several biased building choices.

\xhdr{Investigation of Real-World Pipelines}
The algorithmic fairness literature has a dearth of knowledge about what on-the-ground pipelines look like. Although many papers outline abstracted pipelines (including our own), the actual model creation steps taken by practitioners for a variety of real-world systems (e.g. in consumer finance, healthcare, hiring systems) are at the very least not cataloged in any centralized place. Bias entering from choices along the pipeline cannot be studied, measured, and mitigated if we do not know what choices are being made---thus mapping out actual pipeline choices is a necessary step to operationalize a pipeline-aware approach to fairness.\\
\textbf{Action Items:}  (1) We encourage exploration and documentation of pipeline decisions made in a variety of machine learning pipelines in different applications; and (2) centralization of this information so that researchers can study how pipeline decisions they may have been unaware of impact fairness behavior.

\xhdr{Disconnect between Problem Identification and Mitigation}
Bias problems are often discovered in disciplines on the fringe or outside of machine learning, such as human-computer interaction literature and database management~\cite{sambasivan2021everyone}. 
Since many papers which point to problems along the AI pipeline, especially in the earlier stages, are structured around interviews---they may elide more low-level technical sources of pipeline choices leading to bias, e.g. such as record linkage issues leading to non-IID data dropping. 
While they are helpful in the important first step of identifying problems, they provide little direction for creating operationalizable solutions to these problems, or often even sufficiently detailed information about each system studied to be able to understand the mapping between data collection problems and model behavior. 

Symmetrically, many papers in the mainstream algorithmic fairness literature do not test their techniques on real decision-making systems–or even on datasets beyond Adult, German Credit, and a few others. Though case studies may often be disregarded as implementing old methods and thus not novel, it is crucial that we give them more attention so that we learn about the failure modes of the techniques that we create. While there are papers about the overall problems production teams face when implementing fairness goals, e.g. ~\cite{richardson2021towards, madaio2020co,holstein2019improving,sambasivan2021everyone}, these do not provide an in-depth catalogue of the successes and failures of all pipeline intervention techniques in practice. Proposed methods to intervene on the machine learning pipeline should be tested in a variety of real-world systems to see where they fail and succeed. \\
\textbf{Action Items:}  (1) We encourage bridging the problems identified by the HCI, CSCW, and other literatures, with the technical detail of the algorithmic fairness literature to introduce mitigation techniques for data collection harms; (2) In particular, we encourage AI Fairness researchers to build off of tools for addressing generalized data and modeling pipeline issues, i.e. not specialized to fairness problems, and adapt them for debiasing ML systems--for example, extending Breck et al.~\cite{breck2019data}'s data validation pipeline to address fairness concerns; 
(3) We encourage testing the pipeline-based bias mitigation methods proposed to date in on-the-ground ML systems. This is necessary to uncover which perform best under various circumstances, determine  failure modes, and learn how to integrate these techniques into actual ML systems.

\xhdr{Interaction Effects}
Most fairness-related papers focus on \emph{one issue} in the pipeline: they present a bias problem, then often give a mitigation method, implicitly assuming that the identified source of bias is the only one of interest to the model practitioner, and the suggested intervention will not lead to other effects on the model, including other forms of bias. That is, much of the literature fails to provide insight into how different biases and mitigation techniques along the pipeline interact. Does solving each issue in isolation work, or does a pipeline-aware approach to fairness need to engage with interaction effects to work in practice? There are a few papers that show the impacts of the intersection between multiple sources of bias has on the effectiveness of interventions--e.g., Li et al.~\cite{li2022more} point to how active sampling techniques to address sampling bias can make models \emph{more} unfair when label bias is present. However, any set of choices on the machine learning pipeline may have interaction effects, suggesting many paths for exploration.
Beyond studying interaction effects, there are very few papers (with some exceptions~\cite{10.1145/3531146.3533132,10.1145/3351095.3372874}) 
which engage with the entire pipeline: taking fairness into account when making every modeling decision, and testing along the way---which is the end goal of a pipeline-based approach to fairness. \\
\textbf{Action Items:} We recommend (1) investigating interaction effects of various sources of bias, \textit{and} bias mitigation strategies across the ML pipeline, and (2) developing tools that allow for such exploration (for example, extending~\citep{akpinar2022sandbox}.)

\xhdr{Holes in Research: Measurement Methods and Others}
Within the few papers that do discuss pipeline-aware approaches to fairness, most are problem identification or mitigation techniques— only 17\%\footnote{We calculate this by taking all of the measurement papers identified (67) and subtracting the number that simply introduce new fairness metrics (15), and dividing this number (52) by the total number of pipeline fairness papers identified.} of the papers that we identified are actually providing techniques to measure whether or not a given pipeline choice will lead to downstream unfairness ex-ante. But measurement is key because it may allow us to decide when or when not to take an action. How can we effectively measure algorithmic harms? Relatedly, there are many proposed solutions of how to solve a problem once it has already happened--e.g. unrepresentative data, etc.---but how can we develop tests to \emph{prevent} choices that lead to algorithmic harms? For example, can we predict beforehand whether and how fairness problems will result from the way in which a model is integrated into a given decision structure? 
Additionally, we find that research on \emph{how} bias can enter machine learning decision systems via choices made along the pipeline is unevenly distributed across the pipeline. There are some areas of the pipeline that are well-studied in this regard, such as data collection; but the majority of the pipeline has light-to-no coverage: e.g. viability assessments, problem formulation, and large parts of organizational integration. These areas must be prioritized for searching for fairness failure modes and mitigation techniques.
Finally, the majority of the research to date measures the effect of various pipeline choices on common definitions, e.g. demographic disparity,  accuracy disparity, etc. How can we map the pipeline in a general enough fashion that we may be able to understand how more contextualized notions of fairness are impacted by pipeline decisions? \\
\textbf{Action Items} (1) We encourage building methods for \emph{measuring} the harms introduced by decisions along the machine learning pipeline; (2) Exploring pipeline decisions that have received little attention; and (3) exploring the fairness effects of pipeline decisions beyond the most common metrics. \vspace{0.2em}

\xhdr{Guidance on Choosing Among Imperfect Design Alternatives}
ML practitioners often face choices between imperfect/biased alternatives.
It will almost always be the case that with adequate effort, they can produce fairer, but not perfectly fair models. Any choice a designer makes will likely lead to some form of bias, some more and some less problematic in the given decision-making domain. However, there is little guidance in the literature on deciding when one alternative is better than another in light of all contextual considerations, when certain biases are conditionally acceptable, and how the answers to these questions depend on the context. \\
\textbf{Action Items}: (1) Developing normative guidelines to weigh a variety of imperfect/biased design choices against each other is a crucial avenue for collaboration between ML experts and ethicists; (2) Exploring how different types of bias are currently traded off in practice when they are discovered; and (3) Understanding if there are any relationships or couplings of different sources or forms of bias that often go together, or have opposing relationships. 

\section{Pipeline Cards}
\label{app:cards}

Here, we introduce \emph{pipeline cards}, a framework for interrogating the machine learning development pipeline in order to search for and mitigate sources of bias.

\subsection{Viability Assessment
}

\subsubsection{\underline{Fit:} Is a machine learning system a good solution for the problem?}
\begin{itemize}
    \item What are the policy or business goal(s)?
    \begin{itemize} 
        \item What tensions or trade-offs might exist between these goals?
    \end{itemize}
    \item How can introducing an ML model promote that goal?
    \begin{itemize}
        \item What data is accessible for the project– is there a sufficiently good proxy available for the desired prediction task? ~\cite{raji2022fallacy}
    \end{itemize}
    \item What alternatives are there to introducing a new ML system? What are the benefits and downsides of these other approaches when compared to making an ML system?
\end{itemize}

\subsubsection{\underline{Capacity:} Can the organization build and maintain the system?}
\begin{itemize}
    \item Is there organizational capacity to build and maintain this algorithm (data, expertise, budget)?
    \item Does the project have community and organizational buy-in? What is the evidence to that effect?
\end{itemize}

\subsection{Problem Formulation}

\subsubsection{Prediction Task}

\begin{itemize}
    \item What is my numerical proxy that  will be used for prediction? How does it relate to the policy goal?

    \begin{itemize}
        \item Do I have any alternatives? Why have I chosen the one I did?
        
        \item Is my proxy valid?~\cite{coston2022validity} Some ways to test:
        \begin{itemize}
            \item Is the proxy measuring a facially different outcome or behavior (e.g. arrests instead of convictions in a pretrial risk assessment)?
            \item Is there ground truth data about the behavior or outcome that the model is attempting to predict? Does the proxy correlate well with that information? 
            \item If there are multiple possible prediction proxies available, are they all highly correlated?
        \end{itemize}

        \item Is the proxy equally viable and closely related for the entire population the model will be used over?

    \end{itemize}
\end{itemize}

\subsubsection{Input Features}

\begin{itemize}
    \item What input features are being included to predict the proxy?
    \begin{itemize}
        \item Is there reason to believe some of these input features will be of varying quality across different demographic populations?
    \end{itemize}
\end{itemize}

\subsubsection{Problem Universe}
\begin{itemize}
    \item Is this task being performed on the correct population in order to solve the policy goal? (E.g. is predicting likelihood of dropout for educational intervention better to do on 10th grade or 11th grade students?) 
    \item Are some groups disproportionately under-represented in the universe definition relative to others?
\end{itemize}

\subsection{Data Collection}

\subsubsection{General}

\begin{itemize}
    \item What is the relevant population for the project (demographically, temporally, geographically)? Is there access to data over that entire population?
    \begin{itemize}
        \item How might some individuals or groups be left out of this data?
    \end{itemize}

    \item Does the data have to be i.i.d. (randomly sampled)? How far from a random sample is the data being collected or used?
    \begin{itemize}
        \item E.g. Is the data only available based on condition? (e.g., only those who filled out information release form when applying for loan can be a part of dataset)
    \end{itemize}

\end{itemize}

\subsubsection{Collecting Own Data}

\begin{itemize}
    \item Refer to Datasheets for Datasets~\citep{gebru2018datasheets}.
\end{itemize}

\subsubsection{Using available data}

\begin{itemize}
    \item Does the data  have information that is likely to be of higher quality for some groups (e.g. home address?)
    \item What are the potential mechanisms that could lead to divergent data quality across groups? 
    \item How is data being linked across datasets?
    \begin{itemize}
        \item Are the processes for matching individuals across data sources equally accurate across different populations? 
        \begin{itemize}
            \item For instance, married vs maiden names may bias match rates against women, while inconsistencies in handling of multi-part last names may make matching less reliable for hispanic individuals.
        \end{itemize}
        \item A data loading process that drops records with “special characters” might inadvertently exclude names with accents or tildes.

    \end{itemize}

\end{itemize}

\subsection{Data Preprocessing}

\begin{itemize}
    \item How is missing or damaged data being handled?
    \begin{itemize}
        \item Are some groups affected by these data issues more than others?
        \item Data be imputed instead? If so, how might this process impact different demographic groups differently?
    \end{itemize}
    \item What data transformations are being used? Might they have different effects or accuracy across different groups?
    \begin{itemize}
        \item E.g., Using data transformations with differential error rates across groups (e.g., using word embeddings trained on an English corpus for user-generated text that may contain submissions in other languages)
    \end{itemize}
    \item How are features built from the data? How are they encoded?
    \begin{itemize}
        \item Might some of these definitions or encodings have differential meaning across groups (e.g., distance-based features relative to a home address that is less likely to be stable/current for more vulnerable populations)
    \end{itemize}
\end{itemize}

\subsection{Statistical Modeling}

\begin{itemize}
    \item What model type (hypothesis class)?
    \begin{itemize}
        \item How might it impact different demographic groups differently—e.g. choosing a learning rule that over- or under-emphasizes outliers or minority populations (e.g. differentially private learning rules under-emphasize minority populations~\cite{bagdasaryan2019differential})
    \end{itemize}
    \item What is the learning rule? Might it have an inductive bias that benefits or harms a portion of the data population?
    \item What is the loss function and how well does it reflect the project’s goals? Should any additional constraints or incentives be added?

\end{itemize}

\subsection{Testing and Validation}

\subsubsection{Metrics}

\begin{itemize}
    \item What metrics determine “model performance” in the pipeline? How is one model chosen over another?
    \begin{itemize}
        \item Is any notion of fairness being considered? (E.g. even differential accuracy?) \item How are hyperparameters chosen– can behaviors beyond accuracy be included as a tiebreaker?
        \item How is the robustness of these metrics being ensured? Am I performing trials, or using cross-validation?

    \end{itemize}
\end{itemize}

\subsubsection{Representative Testing Data}

\begin{itemize}
    \item Will performance be tested on data that is from the same distribution as my deployment population? (e.g. is a credit risk assessment system tested on US data to be used in the UK?)
\end{itemize}

\subsubsection{Representative Testing Environment}

\begin{itemize}
    \item In addition to model-only testing, will the model be tested in the full decision system into which it will be embedded (i.e. with human components as well)?
\end{itemize}

\subsection{Deployment and Monitoring}

\subsubsection{Deployment}

\begin{itemize}
    \item How will the model be used as a component of the decision system into which it is embedded?
    \begin{itemize}
        \item Will the model’s predictions directly become the final decision? 
        \item If there is human involvement, where and how will that occur?
        \item How much discretion do humans have over adhering to model recommendations? 
        \item How are model predictions communicated to decision-makers? Is there reason to believe that the way results are communicated will result in bias towards a particular group?
        \item Is there potential for disparities arising from the downstream actions the model is informing (e.g., english-only outreach calls, after school programs that might not be accessible to students with family obligations, etc)?

    \end{itemize}
\end{itemize}

\subsubsection{Monitoring}

\begin{itemize}
    \item Will my system be monitored? What conditions will it monitor (accuracy, fairness)?
    \begin{itemize}
        \item How will errors or issues identified by this monitoring be integrated into system updates or retraining?
    \end{itemize}
    \item Under what conditions will the monitoring system trigger a warning, or even shut the system down? What are the organizational processes for these warnings and/or shutdowns?
    \item What avenues do affected individuals have to challenge decisions or identify/correct errant data?

\end{itemize}

\subsection{Other Aspects of the Pipeline}

\begin{itemize}
    \item What other design decision points exist in my pipeline? What are the decisions I’ve made at those junctures? How might they affect the model’s fairness behavior?
\end{itemize}

\end{document}